\documentclass[runningheads]{llncs}
\usepackage[T1]{fontenc}
\usepackage{graphicx}
\usepackage{booktabs}
\usepackage[misc]{ifsym}

\usepackage{caption}
\usepackage{subcaption}
\usepackage{booktabs} 
\usepackage{tabularx}

\usepackage[capitalize]{cleveref}
\crefname{section}{Sec.}{Secs.}
\Crefname{section}{Section}{Sections}
\Crefname{table}{Table}{Tables}
\crefname{table}{Tab.}{Tabs.}

\usepackage{xcolor}
\usepackage{colortbl}
\usepackage{listings}
\definecolor{green}{RGB}{113,165,55}
\definecolor{blue}{RGB}{1,158,213}
\definecolor{red}{RGB}{183,38,25}
\definecolor{LightGreen}{HTML}{d4edda}  % 定义淡绿色
\definecolor{LightBlue}{HTML}{d1ecf1}   % 定义淡蓝色

\usepackage{multirow}

\newcommand{\equalcontrib}{\thanks{Both authors contributed equally to this work.}}
\newcommand{\corr}{(\Letter)}
\usepackage{mwe}
\begin{document}

\title{SaccadeDet: A Novel Dual-Stage Architecture for Rapid and Accurate Detection in Gigapixel Images}

\toctitle{SaccadeDet: A Novel Dual-Stage Architecture for Rapid and Accurate Detection in Gigapixel Images}

\titlerunning{SaccadeDet for Gigapixel Detection}
% If the full title of your paper is short enough to also fit in the running head, you can omit the abbreviated paper title here. You can check as follows: if you comment out the \titlerunning line, something will appear in the header of all odd-numbered pages of your PDF from page 3 onward. This something is either the full title (in which case all is well), or the error message "Title Suppressed Due to Excessive Length". If this error message appears, you're going to want to provide an abbreviated title within the \titlerunning command, because if you won't do it, Springer will do it for you.

%N.B.: Author information (both in the \author{} and \authorrunning{} command) should only be present in the Camera-Ready Version of your paper. The version that you initially submit for review, ought to be double-blind. So, when initially submitting your paper, use:
% \author{Author information scrubbed for double-blind reviewing}
\author{Wenxi Li\inst{1}\equalcontrib \and
Ruxin Zhang\inst{2}{*} \and
Haozhe Lin\inst{3} \and
Yuchen Guo\inst{3} \thanks{Corresponding author: Yuchen Guo.} \corr \and
Chao Ma\inst{1} \and
Xiaokang Yang\inst{1} }

\tocauthor{Wenxi Li, Ruxin Zhang, Haozhe Lin, Yuchen Guo, Chao Ma, 
Xiaokang Yang}
% Haozhe Lin\inst{1} \and
% Calvin Cordozar Broadus Jr.\inst{2,3}  \and
% Antwan Andr\'e Patton\inst{1}\orcidID{0000-1111-2222-3333}
% You may leave out the orcidID information, if you want to.
% Use \corr to indicate the corresponding author. Note the spacing around the \corr command. Only one author can be the corresponding author.

%N.B.: comment out the \authorrunning{} command for the double-blind version of your paper submitted for review. Later, if your paper is accepted, use the command for the Camera-Ready Version.
\authorrunning{W. Li et al.}
% First names are abbreviated in the running head.
% If there is one author, write 'A.L. Benjamin'.
% If there are two authors, write 'A.L. Benjamin and C.C. Broadus Jr.'
% If there are more than two authors, '[...] et al.' is used.

% \institute{Paper ID 399}
% \institute{
% Beijing National Research Center for Information Science and Technology, \\ Tsinghua University, Beijing, China\\
% \email{linhz@mail.tsinghua.edu.cn,yuchen.w.guo@gmail.com}
% \and 
% MoE Key Lab of Artificial Intelligence, AI Institute, \\ Shanghai Jiao Tong University, Shanghai, China  
% \email{\{wenxi.li,chaoma,xkyang\}@sjtu.edu.cn}
% \and
% Tsinghua Shenzhen International Graduate School, \\Tsinghua University, Shenzhen, China\\
% \email{zrx18@mails.tsinghua.edu.cn}
% % \and
% % Fictional West Coast University, Long Beach CA 90840, USA \email{ccb@fwcu.fake}
% % \and
% % Secondary European Affiliation, Tiergartenstr. 17, 69121 Heidelberg, Germany
% }

\institute{
MoE Key Lab of Artificial Intelligence, AI Institute, \\ Shanghai Jiao Tong University, Shanghai, China  
\email{\{wenxi.li,chaoma,xkyang\}@sjtu.edu.cn}
\and
Tsinghua Shenzhen International Graduate School, \\Tsinghua University, Shenzhen, China\\
\email{zrx18@mails.tsinghua.edu.cn}
\and
Beijing National Research Center for Information Science and Technology, \\ Tsinghua University, Beijing, China\\
\email{linhz@mail.tsinghua.edu.cn,yuchen.w.guo@gmail.com}
% \and
% Fictional West Coast University, Long Beach CA 90840, USA \email{ccb@fwcu.fake}
% \and
% Secondary European Affiliation, Tiergartenstr. 17, 69121 Heidelberg, Germany
}

\maketitle              % typeset the header of the contribution

\begin{abstract}
   The advancement of deep learning in object detection has predominantly focused on megapixel images, leaving a critical gap in efficient processing of gigapixel images. These super high-resolution images present unique challenges due to their immense size and computational demands. To address this, we introduce `SaccadeDet', an innovative architecture for gigapixel-level object detection, inspired by the human eye saccadic movement. The cornerstone of SaccadeDet is its ability to strategically select and process image regions, dramatically reducing computational load. This is achieved through a two-stage process: the `saccade' stage, which identifies regions of probable interest, and the `gaze' stage, which refines detection in these targeted areas. Our approach, evaluated on the PANDA dataset, not only achieves a 8$\times$ speed increase over the state-of-the-art methods but also demonstrates significant potential in gigapixel-level pathology analysis through its application to Whole Slide Imaging.

\end{abstract}

\section{Introduction}
\label{sec:intro}

\begin{figure}
  \centering
  \begin{subfigure}{1.0\linewidth}
  \centering
    \includegraphics[height=26mm]{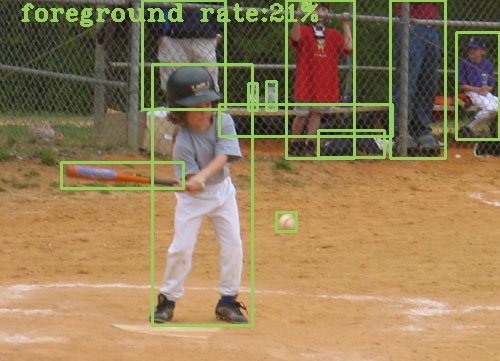}
    \includegraphics[height=26mm]{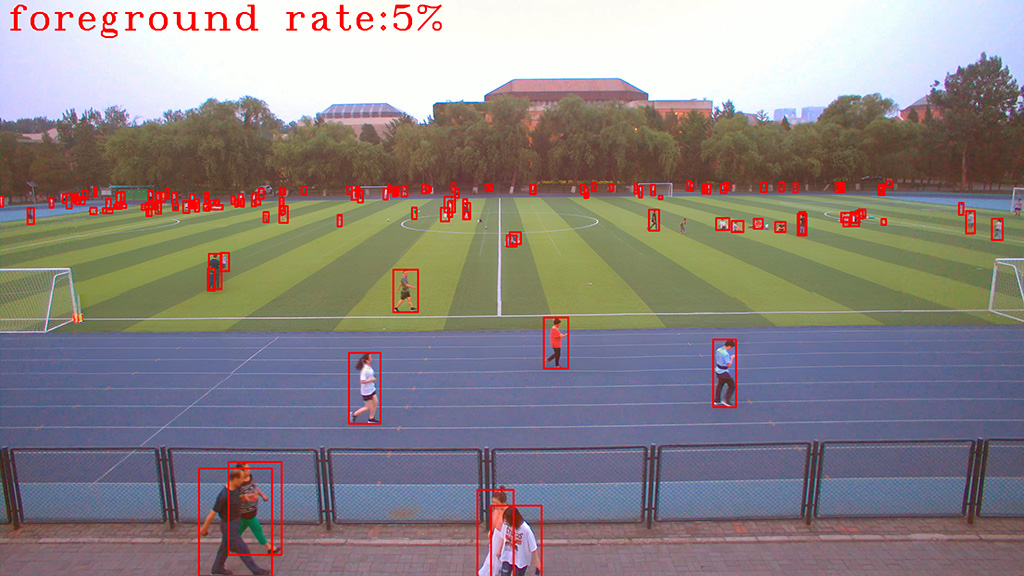}
    \caption{Comparison of foreground rates for COCO~\cite{lin2014microsoft} and  PANDA~\cite{wang2020panda}.}
    \label{fig:issue_background}
  \end{subfigure}
  \\
  \begin{subfigure}{1.0\linewidth}
    \centering
    \includegraphics[width=88mm]{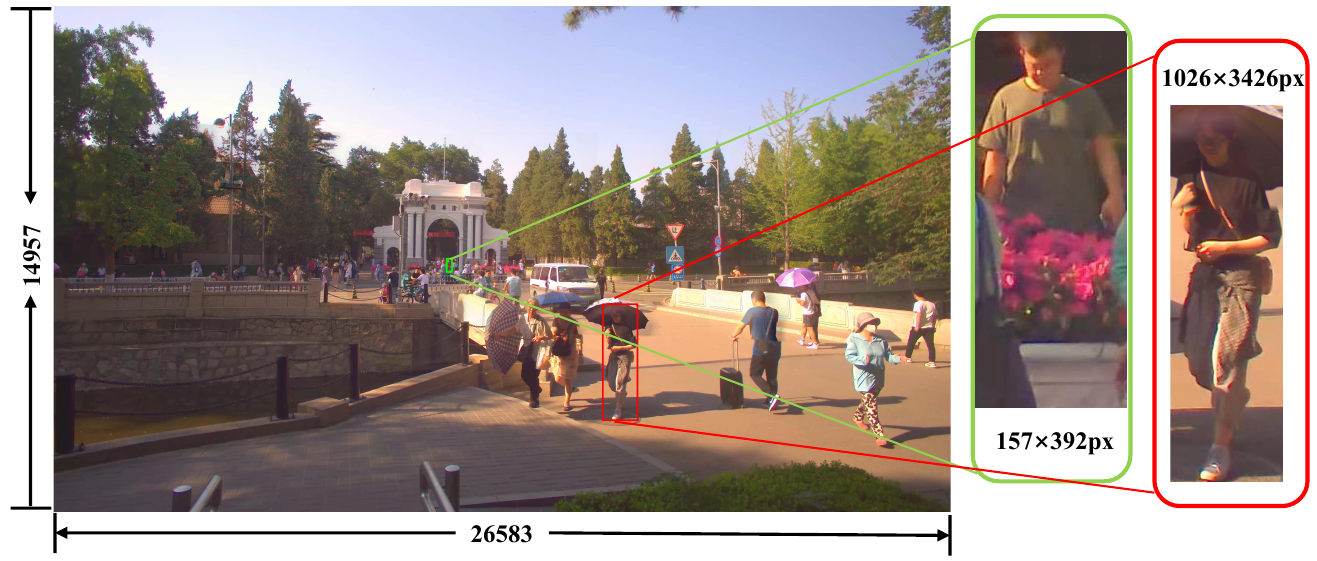}
    \caption{Large-scale variation in PANDA~\cite{wang2020panda}.}
    \label{fig:issue_scale}
  \end{subfigure}
  \caption{\textbf{Two distinct characteristics of gigapixel images.} (a) Wide-field gigapixel images exhibit a higher background rate compared to megapixel images. (b) Images showcasing objects with over 100$\times$ scale variation, demonstrating the extreme size disparities within a single image.}
  \label{fig:issue}
\end{figure}

Object detection, a critical field in computer vision, has seen extensive research and significant advancements, particularly in the megapixel domain~(thanks to MS COCO~\cite{lin2014microsoft}). However, the shift towards super high-resolution gigapixel images~\cite{fan2022speed,Lin2024gigatraj,liu2024gigahumandet,ma2024grounding,wang2020panda,zhang2023skimming}, especially prevalent in medical and surveillance applications, introduces a pressing challenge: the dramatic slowdown of existing detection methods. This slowdown is primarily due to the super high resolution in gigapixel images, which overwhelms traditional detection methods. This phenomenon is coming from two aspects: the extensive background pixels in the wide field of view as shown in \cref{fig:issue_background}, which hinder detection speed and accuracy, and the large scale changes within each image as shown in \cref{fig:issue_scale}, which the receptive field of traditional detectors struggle to cover, leading to compromised accuracy.

\begin{figure}[t]
  \centering
  % \fbox{\rule{0pt}{2in} \rule{0.9\linewidth}{0pt}}
   \includegraphics[width=1.0\linewidth]{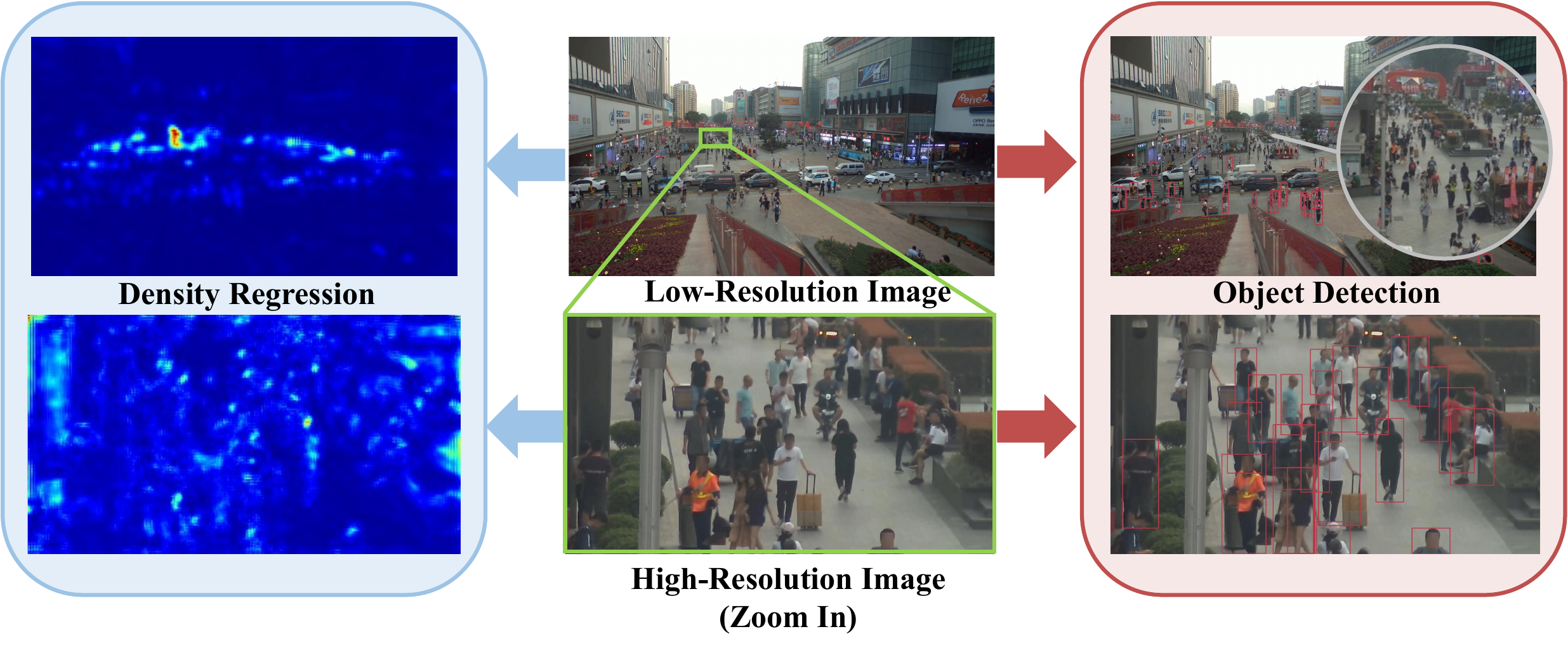}
    % \vspace{-0.4cm}
    \caption{\textbf{Comparative analysis of low-/high-resolution images in density regression and object detection.} 
    We employ zoom-in to differentiate between low- and high-resolution. It demonstrates that density regression~(CSRNet~\cite{li2018csrnet}) excels in coarse-grained localization using low-resolution images, whereas object detection~(RTMDet~\cite{lyu2022rtmdet}) achieves superior fine-grained detection with high-resolution images.
}
   % \vspace{-0.4cm}
   \label{fig:current_methods}
\end{figure}

The central challenge in processing gigapixel images is optimizing for speed without sacrificing accuracy. Previous methods~\cite{fan2022speed,liu2024gigahumandet,zhang2023skimming} have made some attempts at gigapixels, however, the speed is still lower than 1 FPS, which is far from real-time.
% An obvious solution is to use small object detection based on the low-resolution images. 
Significant advancements have been achieved in small object detection, enabling the identification of objects in low-resolution images. However, these methods still fall short in accurately locating objects in distant regions of gigapixel images, leading to numerous missed detections as shown in \cref{fig:current_methods}. In typical crowd counting methods~\cite{li2021learning,li2018csrnet}, although effective at low-resolution image processing, the density maps generated by them typically yield only coarse-grained locational information.

Addressing this challenge, we focus on a critical aspect of gigapixel image processing: the strategic selection of image regions for processing. Our hypothesis is that by intelligently selecting specific regions within these super high-resolution images, we can significantly reduce the computational load, thereby accelerating the detection process. This approach is akin to how the human eye scans a scene, focusing on areas of interest while disregarding irrelevant background information. Gigapixel images often contain vast areas of non-essential data. For instance, in the PANDA dataset~\cite{wang2020panda}, only about 5\% of the image area typically contains relevant information as shown in ~\cref{fig:issue_background}. By concentrating our computation on these critical areas, we can drastically enhance speed.

In this paper, we propose a novel gigapixel detection architecture named SaccadeDet, which embodies the above principle. It comprises two distinct stages: the `saccade' stage, which employs an advanced model to identify potential regions of interest, and the `gaze' stage, which conducts detailed analysis on these selected regions. This dual-stage approach mimics the efficiency of human vision, enabling rapid processing of gigapixel images without compromising on detection accuracy.

A key innovation in our `saccade' stage is the introduction of a multi-scale density estimation method. This technique allows for the rough estimation of different-sized object distributions on low-resolution images, enabling more precise application of post-processing algorithms for patch slicing. This multi-scale approach not only enhances accuracy but also ensures optimal scale for detection to boost inference speed.
Moreover, we employ variably sized windows for slicing images, depending on their scale. Consequently, the super high-resolution images are segmented into patches with different sizes and forwarded to the gaze stage. Previous findings indicate that larger objects can be effectively detected without the need for their original size~\cite{najibi2019autofocus,singh2018sniper}. Therefore, we standardize the size of all patches, resizing them according to a principle that prioritizes high resolution for small objects and low resolution for larger ones.

Our approach diverges significantly from existing gigapixel methods, such as SPDet~\cite{zhang2023skimming} and PAN~\cite{fan2022speed}. 
Unlike these methods, we employ density regression at lower resolutions, which enables our method to outperform them in terms of speed, achieving up to an 8$\times$ improvement. This enhancement in speed is primarily due to the superior parallelism of each cropped patch, which not only significantly boosts inference speed but also maintains high accuracy. Additionally, our application of this method to Whole Slide Imaging demonstrates its versatility and potential impact in various high-resolution imaging fields.

In summary, the main contributions of this paper are threefold:
\begin{itemize}
\item We introduce SaccadeDet, a novel gigapixel detection architecture that effectively mimics human visual processing through its dual-stage approach. This architecture comprises the `saccade' stage for rapid identification of regions of interest and the `gaze' stage for detailed analysis of these selected regions. This methodology allows for the efficient processing of gigapixel images, significantly reducing the computational load without compromising detection accuracy.

\item In the `saccade' stage, we apply a multi-scale density method to spot objects of different sizes at lower resolutions, and adjust image slicing with various window sizes to balance detection accuracy and speed. Then, in the `gaze' stage, we resize these image segments, using high resolution for small objects and low resolution for large ones.

\item Our method uses density regression at lower resolutions and optimizes patch processing with parallel techniques, resulting in an 8$\times$ speed improvement over current gigapixel image methods like SPDet and PAN. This increase in speed, coupled with maintained high accuracy, shows its significant potential in high-resolution imaging fields like medical imaging and surveillance.

\end{itemize}

\section{Related Work}
\label{sec:related}
\noindent
\textbf{Object Detection.} 
Object detectors are mainly divided into one-stage detectors and two-stage detectors.
The main pursuit of two-stage object detection is accuracy, and it frames the detection as a “coarse-to-fine” process.
The representative two-stage object detector is RCNN-series\cite{cai2018cascade,girshick2014rich,girshick2015fast,he2017mask,ren2015faster}.
YOLO\cite{redmon2016you} is the first one-stage detector based on deep learning which focuses on speed.
Its following works try to make more improvements, such as more anchors, better architecture, richer training tricks, and so on, 
to bridge the accuracy gap with the two-stage detectors~\cite{yolox2021,liu2016ssd,redmon2018yolov3}. 
Recent one-stage works are more focused on anchor-free design~\cite{duan2019centernet,lan2020saccadenet,law2018cornernet,tian2019fcos}.
In addition, feature representation plays an important role in two kinds of detector architecture.
The pyramid structure is a common design in modern object detectors enabling detection across a range of scales~\cite{he2016deep,liu2021Swin,simonyan2014very}.
Furthermore, FPN~\cite{lin2017feature} introduces features composed of bottom-up paths and top-down paths and becomes the mainstream design.
There are always some researchers who love to try to tackle the scale-variation problem in different ways.
SNIPER~\cite{singh2018analysis,singh2018sniper} propose the scale normalization to train objects in desired scale range for each scale of an image pyramid.
However, this speed is much slower than FPN, so AutoFocus~\cite{najibi2019autofocus} uses multi-stage detection to prune the image pyramid and improve the speed.
TridentNet~\cite{li2019scale} uses multiple branches with different dilation rates to convolution feature maps on different scales.
% Different branches share the same parameters and only use dilated convolution to tackle scale variation.
Most of the works try to extract richer features on full-size images except AutoFocus.
This makes it difficult to maintain the original performance on the gigapixel level.
\\[6pt]
\noindent
\textbf{Super High-Resolution Object Detection.} PANDA~\cite{wang2020panda} introduces the first gigapixel-level human-centric video dataset. This dataset is specifically designed for large-scale, long-term, and multi-object visual analysis.
Several works have focused on processing gigapixel-level images by subdividing them into manageable 
 patches. PAN~\cite{fan2022speed} introduces an innovative framework that arranges patches into compact canvases, optimizing for faster object detection. GigaDet~\cite{chen2022towards} emphasizes a balance between speed and accuracy, employing a uniform patch selection mechanism. Meanwhile, Remix~\cite{jiang2021flexible} processes ultra-high-resolution images on edge devices, adjusting detection based on crowd count and time constraints, although this can sometimes challenge system stability. GigaHumanDet~\cite{liu2024gigahumandet} employs the corner modeling method to avoid the potential issues of a high degree of freedom in center pinpointing. 
Drawing from the aforementioned research and considering the challenges of gigapixel imagery, we have distilled several key insights: Firstly, resampling objects of varying sizes judiciously can help strike a balance between speed and accuracy. Secondly, minimizing time spent processing background areas is crucial for efficiency. Finally, it is imperative to harness the full potential of the detector without compromise.
\\[6pt]
\noindent
\textbf{Crowd Counting.} Crowd counting aims to estimate the number of persons in low-resolution images. 
Thanks to Lempitsky et al.~\cite{lempitsky2010learning} firstly introduce the density map in counting task, 
recent work based on density regression has better performance than detection.
Zhang et al.~\cite{zhang2016single} find that a fixed Gaussian kernel can not cover different sizes of the person in the scene due to the perspective distortion.
They generate the density map via geometry-adaptive kernels calculated by the distance among the nearest k neighbors.
Jiang et al.~\cite{jiang2020attention} propose ASNet to generate the multiple intermediate density maps and use DANet to focus on their corresponding attention regions.
No fancy network structure, CSRNet~\cite{li2018csrnet} modifies the VGG~\cite{simonyan2014very} with dilated convolution and outperforms previous work.
Unlike the work of a lot of complex designs on the network structure,
Li et al.~\cite{li2021learning} show that a simple scale factor can significantly reduce the error without any other modification.
Previous works introduce a detection module to improve the performance of density estimation\cite{liu2018decidenet,lian2019density}, while we introduce the density module into object detection to improve the detection speed.

\begin{figure*}[t]
  \centering
  % \fbox{\rule{0pt}{2in} \rule{0.9\linewidth}{0pt}}
   \includegraphics[width=1.0\linewidth]{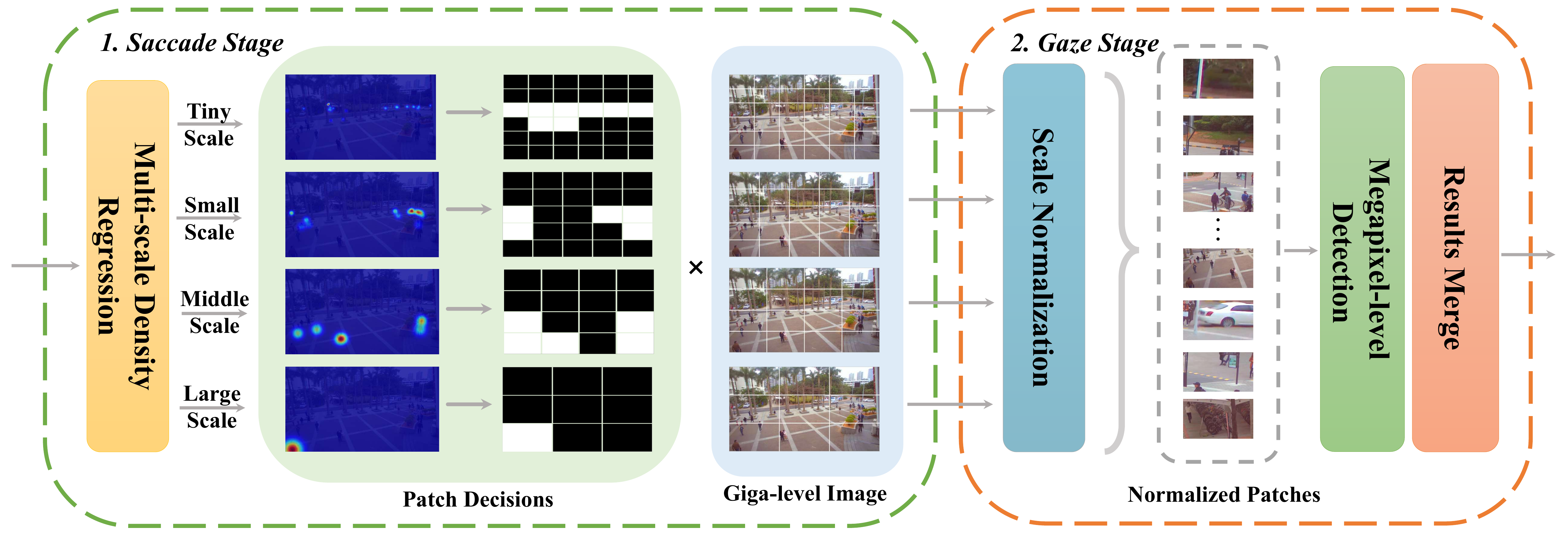}
    % \vspace{-0.4cm}
   \caption{\textbf{An overview of SaccadeDet architecture. } 
%   It contains two stages called saccade and gaze respectively. 
   (1) A low-resolution image is passed to the multi-scale density regression module to generate the density maps. Then, the density maps are divided into grids and calculated the density of objects for each cell.
   The patch with high density is cropped for the gaze stage.
   (2) Multi-scale patches are scaled to the same scale and the megapixel-level detector processes them to output the bounding boxes of each patch.
   Finally, merge all results to generate the bounding boxes of the gigapixel-level image.
   }
   % \vspace{-0.4cm}
   \label{fig:overview}
\end{figure*}

\section{Proposed Method}
\label{sec:method}

\subsection{Preliminaries}

In addressing gigapixel-level object detection, we build upon the foundational concept of feature and image pyramids. The feature pyramid, crucial in handling various object scales as delineated by Lin et al.~\cite{lin2017feature}, necessitates deeper levels for larger scale objects, thereby increasing computational demands. Applying even shallow network structures to gigapixel images quickly surpasses memory and inference time limits. AutoFocus~\cite{najibi2019autofocus}, while a significant step towards real-time gigapixel-level detection, faces challenges with scaling and lacks efficient parallelization in multi-stage detections. Our approach is designed to simultaneously process all scales of the image pyramid, effectively overcoming these challenges.

\subsection{Overall Architecture}
\label{sec:overall_arch}
The current megapixel-level approach has two main weaknesses on gigapixel-level images: 
1) The waste of computation resources on background,
and 2) sliding window cannot tackle the large-scale variation.
We propose a multi-scale density regression to locate the region of interest on low-resolution images, which will drop the most of background area.
To tackle the large-scale variation, we proposed scale normalization to downsample the large objects, which will also be beneficial to speed.
An overview of our architecture is presented in  \cref{fig:overview}.
How the saccade stage and gaze stage work will be shown in the following.

\subsection{Saccade by Multi-scale Density Estimation}
\label{sec:saccade}
In the saccade stage, inspired by the pre-saccadic compression of visual space, we employ a density-based strategy to estimate scale variations across different image regions. This technique allows us to determine the scale by a single assessment, unlike traditional multi-stage zoom-in approaches. We utilize density regression, which requires fewer features and can work with lower resolution images, focusing on predicting object groups rather than individual objects. The approach to generating the density map is informed by existing work in crowd counting, using a dynamic Gaussian kernel to represent object probabilities:

\begin{equation}
  GT_{s}(x)=\sum_{i=1}^{N_{s}}\delta(x-x_i)*G_{\sigma_i}(x),
\end{equation}
where \(\delta(x-x_i)\) signifies the object centers and \(G_{\sigma_i}(x)\) is the Gaussian kernel. 

The samples of the density map are shown in \cref{fig:overview}. $N_{s}$ represents the number of the labeled objects on the $s$-th scale which we will introduce how to divide in the following.
We use a dynamic $\sigma$ here, which is to better represent the scale of the object so that the integral over the whole area of the object is going to be 1.
According to the three-sigma rule of thumb, $\sigma_i$ is calculated by:

\begin{equation}
  \sigma_i=\max (x_{i}.height, x_{i}.width)//3.
\end{equation}
where $x_{i}.height$ and $x_{i}.width$ are the height and width of object $x_{i}$.

Although the dynamic $\sigma$ implies the scale of the object, 
it is still difficult to obtain the scale directly from a single density map.
We leave the challenge of scale variation mainly in the next stage to solve.
Here we roughly divide the density map into four scales based on the size of the object,
which are named $Tiny\ Scale$, $Small\ Scale$, $Middle\ Scale$ and $Large\ Scale$. 
We will discuss the range of different scales for specific datasets in \cref{sec:experiments}.
Our multi-scale density regression will get related density maps from low-resolution images for further processing.

\subsection{Scale-aware Mean Squared Error Loss}
In this subsection, we modified mean squared error loss to address the imbalance of density maps at different scales during training.
Wang et al.~\cite{wang2020panda} have shown the distribution of person scale in multiple datasets and it is easy to see that examples imbalances are widespread in these datasets.
In \cref{sec:saccade}, we introduced our design for the multi-scale density map, which only considers the sizes of the objects.
% The size distribution is not uniform distribution.
The object scale imbalance in the dataset is transformed into density map scale imbalance.
Formally, we add a modulating factor to the vanilla Mean Squared Error~(MSE) loss for different scales.
We define our loss as:

\begin{equation}
  L(\Theta)=\sum_{s=1}^{S}\ \alpha_{s}\left \| \mathcal{F}_{s}(X, \Theta)-GT_{s} \right \|_{2},
\end{equation}
where $s$ represents the scale of the density maps.

In the previous method, the data will be resampled to tackle the imbalance problem. 
In our density estimation, each training sample has example imbalance on four scales, 
so this fine-grained modulating factor is a more advantageous method.

\subsection{Patch Generation}
\label{sec:patch_generation}
Our multi-scale density regression could effectively manage objects of varying sizes. When we slice the super high-resolution image based on density maps, it is critical to ensure that larger objects are not inadvertently segmented, preserving their integrity in the image analysis.

Once the image is segmented, we calculate the integral within each patch. This integral acts as an indicator of the density of persons within that segment. We focus our detection efforts on patches with higher integral values, as they suggest a higher density of containing detectable objects.
 
To ensure that no object is partially missed or overlooked, we have refined our grid-based strategy. Each grid cell is extended by 1.2 times its original size, creating overlaps between adjacent patches. This overlap is a critical aspect of our methodology, ensuring that each object is fully captured, even if it spans multiple grid cells. By integrating this overlap system, we significantly reduce the risk of losing contextual information and improve the detection of objects that may otherwise have been partially or completely missed.

\subsection{Gaze with Scale Normalization}
\label{gaze_stage}
The aim of the gaze stage is to perform object detection on the high density areas obtained during saccades.
The current challenge is the large-scale variation which is a traditional problem of object detection.
Building on prior research~\cite{najibi2019autofocus,singh2018sniper}, we recognize that larger objects can be detected effectively without retaining their original size. Consequently, we adopt a strategy of standardizing the size of all patches, resizing them in a manner that favors high resolution for smaller objects and lower resolution for larger ones.
We formulate the patches generated from the saccade stage with four scales as \{Patch$_{t}$, Patch$_{s}$, Patch$_{m}$, Patch$_{l}$\}.
We define the scale of Patch$_{t}$ as the standard scale, 
so the patch on other scales has sizes that exceed required for detection and could be downsampled for accelerating.
Then all patches have the same scale and can be processed by the same detector.
This design can bring more samples to the training of the detector and is completely parallel in the inference process, which brings new improvements in both speed and accuracy.

\begin{figure*}[t]
  \centering
  % \fbox{\rule{0pt}{2in} \rule{0.9\linewidth}{0pt}}
   \includegraphics[width=1.0\linewidth]{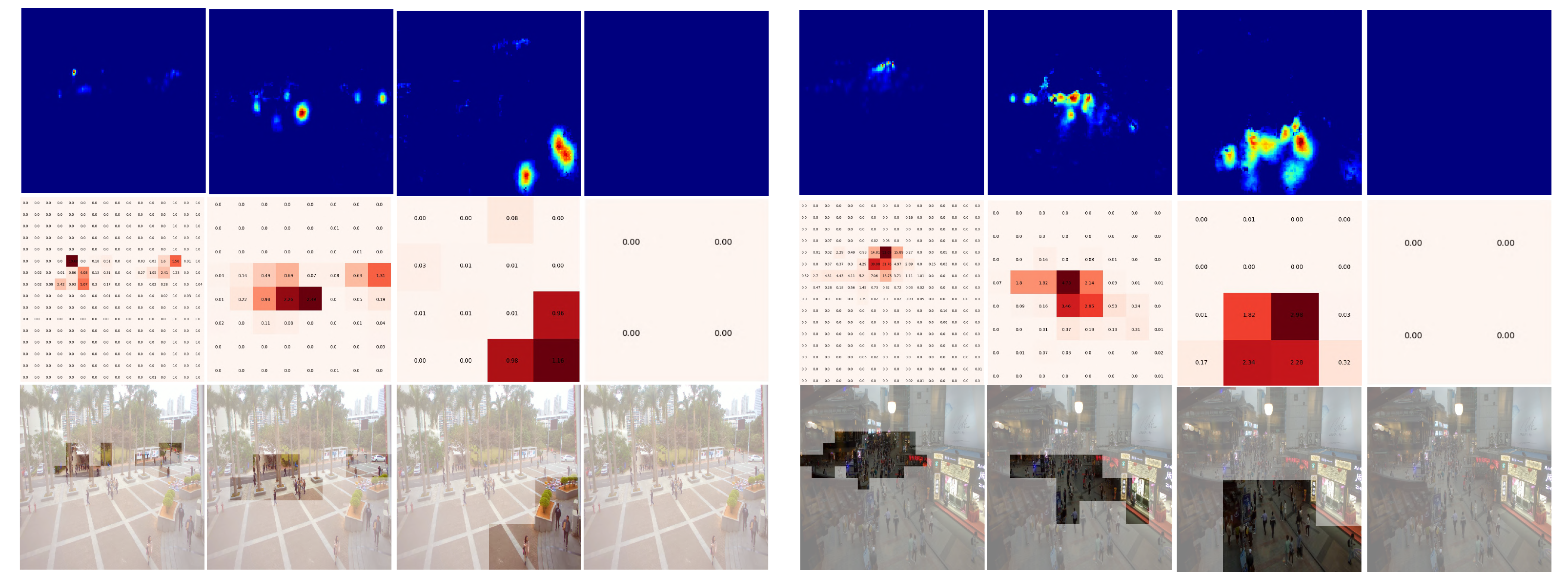}
    % \vspace{-0.5cm}
   \caption{\textbf{Visualization of the procedure in saccade stage.} We show the density map, the object density of the patch and the selected patch. 
   The masks represent the corresponding patches that are discarded.
   It can be seen that most of the background has been discarded. 
   This phenomenon indicates that this stage can provide a more focused patch for the gaze stage.}
   % \vspace{-0.4cm}
   \label{fig:model_analysis}
\end{figure*}

\begin{table}[t]
  \centering
  \setlength\tabcolsep{8pt} % 调整列间距
  % \resizebox{\columnwidth}{!}{ % 调整表格宽度以适应单栏
    \caption{\textbf{Comparisons with the state-of-the-art methods.} Our approach achieves an eightfold increase in speed over SPDet, the current top performer, without compromising on accuracy.}
  \begin{tabular}{@{}l|c|ccc|c@{}}
    \toprule
    Method & AP$_{50}$ & AP$_{S}$ & AP$_{M}$ & AP$_{L}$ & FPS\\
    \midrule
    RetinaNet~\cite{lin2017focal}         & - & 0.221 & 0.561 & 0.740 & 0.1 \\
    FasterRCNN~\cite{ren2015faster}       & -  & 0.190 & 0.552 & 0.744 & 0.07 \\
    YOLOX~\cite{yolox2021}                 & 0.639  & 0.256 & 0.550 & 0.648 & 0.65 \\
    RTMDet~\cite{lyu2022rtmdet}           & 0.658 & 0.240 & 0.548 & 0.696 & 0.85 \\
    ClusDet~\cite{yang2019clustered}      & 0.718  & 0.219 & 0.696 & 0.782 & 0.12  \\
    DMNet~\cite{gao2018dynamic}           & 0.540  & 0.119 & 0.371 & 0.714 & 0.51  \\
    PAN~\cite{fan2022speed}               & 0.713  & 0.262 & 0.715 & 0.767 & 0.23  \\
    SPDet~\cite{zhang2023skimming}        & 0.755  & 0.287 & 0.731 & 0.790 & 0.41  \\
    \rowcolor{LightBlue} 
    Ours                                  & \textbf{0.760} & \textbf{0.340} & \textbf{0.751} & \textbf{0.808} & \textbf{3.2}  \\
    \bottomrule
  \end{tabular}
  % }
  \label{tab:sota}
\end{table}

\section{Experiments}
\label{sec:experiments}

% In this section, we present the experiments.

\textbf{Dataset.} PANDA~\cite{wang2020panda} is the first human-centric gigapixel-level dataset.It contains 18 scenes with over 15, 974.6k bounding boxes annotated. 
Specifically, there are 13 scenes used for training and 5 scenes used for testing.
\\[6pt]
\noindent
\textbf{Evaluation metrics.}  The gigapixel-level detection can be evaluated as a common detection task.
We report the standard COCO metric of AP$_{50}$.
The AP for small~($< 96\times96$), middle~($96\times96-288\times288$) and large~($> 288\times288$) categories are denoted as AP$_S$, AP$_{M}$ and AP$_{L}$.
It is worth noting that although there may be many cropping or resizing operations on the images during the inference,
the final evaluation must be on the original scale.
Evaluation on each patch and then averaging will cause errors, which is not conducive to the consistency of previous and future studies.
Our method also focuses on efficiency, so we evaluate the speed of different methods on a 3090 GPU with 32 GB memory using Frames Per Second~(FPS).

\noindent
\textbf{Implementation Details.} 
Leveraging the scale factor from Li et al.~\cite{li2021learning}, our density regression model, based on ResNet-18, was efficiently trained. The standard density map was scaled up by a factor of 1000 during training for enhanced precision and then scaled back down during evaluation. According to PANDA~\cite{wang2020panda}, we divided the density map into four scales (as detailed in \cref{tab:4scale}), setting the scale factors $\alpha_{t}$, $\alpha_{s}$, $\alpha_{m}$, and $\alpha_{l}$ at 0.01, 0.1, 10, and 100, respectively. This scaling is necessary as the majority of examples were in the tiny scale, with the other scales having fewer training samples. 
For consistency, we scaled the patches of the other three scales to match the tiny scale, enabling our detector to train solely on tiny scale samples.
Additionally, we utilized MMDetection for the implementation of the detectors. The training set for the megapixel-level detector during the gaze stage underwent preprocessing, including cropping the original image according to 16×16, 8×8, and 4×4 grids. Each detection model's data augmentation and training configurations were aligned with the original papers' settings.
\begin{table}[t]
  \centering
    \setlength\tabcolsep{8pt} % 调整列间距
      \caption{\textbf{Coarse-grained scale definition for the saccade stage.}
  The last column indicates the number of bounding boxes of this scale in PANDA.}
  \begin{tabular}{@{}c|c|c@{}}
    \toprule
    \textbf{Scale Level}  & \textbf{Range} & \textbf{Number}\\
    % \toprule
    \hline
    Tiny Scale  & $<$ 800 & 73522\\
    \hline
    Small Scale  & $\ge$ 800 and $<$ 1600 & 7586\\
    \hline
    Middle Scale  &$\ge$ 1600 and $<$ 3200 & 1365\\
    \hline
    Large Scale  & $\ge$ 3200 & 56\\
    \bottomrule
    % \hline
  \end{tabular}
  % \vspace{-0.6cm}
  \label{tab:4scale}
\end{table}

\subsection{Baseline}
% \vspace{-1.5mm}
For the main results and ablation study, we consider eight typical object detection frameworks: RetinaNet~\cite{lin2017focal}, Faster R-CNN~\cite{ren2015faster}, YOLOX~\cite{yolox2021}, and RTMDet~\cite{lyu2022rtmdet}, ClusDet~\cite{yang2019clustered} , DMNet~\cite{gao2018dynamic}, PAN~\cite{fan2022speed} and SPDet~\cite{zhang2023skimming}.
% The backbone includes Transformer-based architecture~\cite{liu2021Swin} and ConvNets~\cite{he2016deep, sandler2018mobi}.
Since all backbones have the feature pyramid structure, the naive method sliding window~(SW) and the image pyramid based method AutoFocus~\cite{najibi2019autofocus}~(Zoom) are mainly compared in the ablation study.
\\[6pt]
\noindent
\textbf{Sliding Window.} 
% Sliding window~(SW) is a traditional approach. 
We crop the high-resolution images into local patches and infer those patches sequentially.
In our experiments, we define two types of sliding windows to transform the original gigapixel-level images to the megapixel-level patches.
Then, we apply current megapixel-level detectors on each patch and get the results of the local images.
Finally, we will combine the results of each patch to get final detection results.
Two types of sliding windows divide the original image into 16$\times $16 grid and 8$\times $8 grid respectively.
It is worth noting that because the expansion factor could bring better performance, we also use this strategy here for a fair comparison.
\\[6pt]
\noindent
\textbf{AutoFocus.}  AutoFocus~\cite{najibi2019autofocus} represents a coarse-fine approach for efficient multi-scale inference. Implemented on the PANDA dataset~\cite{wang2020panda}, it was adapted from its original COCO~\cite{lin2014microsoft} settings. Our implementation involved a single zoom-in stage, selecting areas for refinement based on the bounding box size with an empirical threshold set at 0.07. The bounding boxes were enlarged to increase contextual information, with sides set to four times the length of the longest side of the previous box.

% \textbf{Sliding Window.} Sliding window is a traditional approach. 
% The traditional sliding window approach involves cropping high-resolution images into smaller patches for sequential processing. We tested two grid sizes, 16×16 and 8×8, to transform gigapixel images into megapixel-level patches. Each patch was processed using current detectors, and the results were combined to generate the final detection output. 
% % Notably, an expansion factor was applied to improve performance and ensure a fair comparison.
% \\[6pt]
% \noindent
% \textbf{AutoFocus.}  AutoFocus~\cite{najibi2019autofocus} represents a coarse-fine approach for efficient multi-scale inference. Implemented on the PANDA dataset~\cite{wang2020panda}, it was adapted from its original COCO~\cite{lin2014microsoft} settings. Our implementation involved a single zoom-in stage, selecting areas for refinement based on the bounding box size with an empirical threshold set at 0.07. The bounding boxes were enlarged to increase contextual information, with sides set to four times the length of the longest side of the previous box.

\subsection{Module Analysis}
To further investigate the behavior of the saccade stage, we visualize the density map, object density of the grid, and the selected grid in \cref{fig:model_analysis}.
We can see that the density map can clearly show the distribution of the crowd.
Then we divide the density map into grids and integrate the inside of each grid to get the density of the existence of persons.
By this time, most of the grid cells with no person have got zero density, and we can use a more suitable threshold to further improve the performance.
Finally, we can choose the cells that need attention based on this result.
The visualization results show the excellent performance of our method.
Our multi-scale density regression can regress the different scale persons to the most suitable density map. 
The patch generation can also gaze at the valuable region for further processing.
% The expansion factor is not shown in our visualization, we will verify it through experiments in the following.

\subsection{Comparisons with the state-of-the-art methods}
We conducted evaluations of state-of-the-art detectors on the PANDA~\cite{wang2020panda}. Prominent among these are RetinaNet~\cite{lin2017focal}, Faster RCNN~\cite{ren2015faster}, YOLOX~\cite{yolox2021}, and RTMDet~\cite{lyu2022rtmdet}, which represent the current mainstream in object detection. Due to limitations posed by CUDA memory capacity, a sliding window method was implemented to facilitate the detection process with these detectors. However, these detectors exhibited low accuracy and processing speed.
In addition, our evaluation included detectors specifically engineered for high-resolution imagery, such as ClusDet~\cite{yang2019clustered} and DMNet~\cite{gao2018dynamic}, and those tailored for super high-resolution image analysis, like PAN~\cite{fan2022speed} and SPDet~\cite{zhang2023skimming}. Despite their advanced design for handling high-resolution inputs, they only achieved a maximum frame rate of 0.41 FPS, significantly below the threshold required for real-time application scenarios.

\begin{table}[t]
  \centering
    \caption{\textbf{Ablation experiments on density boundaries.} The threshold was set at 0.2 for subsequent studies, as this value offered the optimal balance in our analysis. It is worth noting that density can be a number greater than 1, which means there are objects here greater than 1.}
  \begin{tabularx}{\columnwidth}{c|XXXX|c}
    \hline
    Density & AP$_{50}$ & AP$_{S}$ & AP$_{M}$ & AP$_{L}$ & FPS \\
    \hline
    0     & 0.757 & 0.340 & 0.750 & 0.796 & 3.0 \\
    \rowcolor{LightBlue} 
    0.2   & \textbf{0.760}  & \textbf{0.341} & \textbf{0.751} & \textbf{0.808} & 3.2 \\
    0.4  & 0.759  &0.340 & \textbf{0.751} & 0.807 & 3.3 \\
    0.6  & 0.754 & \textbf{0.341} & 0.750 & 0.803 & 3.4 \\
    0.8  & 0.746  & 0.340 & \textbf{0.751} & 0.796 & 3.6 \\
    1.0  & 0.737  & \textbf{0.341} & 0.750 & 0.789 & 3.8 \\
    \hline
  \end{tabularx}
  \label{tab:threshold}
\end{table}

\begin{table}[t]
  \centering
    \setlength\tabcolsep{6pt} % 调整列间距
  % \scalebox{0.8}{
      \caption{\textbf{Ablation Studies for Different Inference Methods:} $\dagger$Note that both AutoFocus and our method dynamically extract patches, leading to variability in processing time and the number of patch proposals per image. The values presented here are averaged for clarity.}
    \begin{tabular}{@{}l|c|c|c|ccc|c@{}}
      \hline
      \toprule
      
        & Strategy & Patch  & AP$_{50}$  & AP$_{S}$ & AP$_{M}$ & AP$_{L}$ &FPS\\
      \midrule
      RetinaN~\cite{lin2017focal} & SW & 256    & 0.732  & 0.303 & 0.751  & 0.688  & 0.07\\
      RetinaN~\cite{lin2017focal}  & SW  & 64    & 0.684 & 0.254 & 0.582  & 0.687  & 0.28\\
      RetinaN~\cite{lin2017focal}    & Zoom    & 98$\dagger$   & 0.681  & 0.290 & 0.698  & 0.692  & 0.23\\
      RetinaN~\cite{lin2017focal}   & Ours         & 42$\dagger$    & 0.750  & 0.300 & 0.756  & 0.716  & 0.43\\
      \hline
      FasterR~\cite{ren2015faster}  & SW & 256  & 0.727  & 0.318 & 0.742 & 0.723 & 0.08\\
      FasterR~\cite{ren2015faster}  & SW  & 64  & 0.684  & 0.240 & 0.610 & 0.702 & 0.30 \\
      FasterR~\cite{ren2015faster}  & Zoom    & 80$\dagger$  & 0.694 & 0.240 & 0.632 & 0.722 & 0.30\\
      FasterR~\cite{ren2015faster}  & Ours      & 42$\dagger$  & 0.752  & 0.308 & 0.757 & 0.760 & 0.45\\
                                  \hline
      YOLOX~\cite{yolox2021}  & SW  & 256   & 0.652  & 0.283 & 0.571 & 0.618 & 0.19 \\
      YOLOX~\cite{yolox2021}  & SW   & 64   & 0.639  & 0.256 & 0.550 &  0.648 & 0.65\\
      YOLOX~\cite{yolox2021}  & Zoom     & 54$\dagger$   & 0.609  & 0.275 & 0.589 & 0.585 & 0.27\\
      YOLOX~\cite{yolox2021} & Ours          & 42$\dagger$    & 0.739  & 0.326 & 0.730 & 0.787 & 1.00\\
                                  \hline
      RTMDet~\cite{lyu2022rtmdet}& SW  & 256   & 0.737& 0.324 & 0.747 & 0.777  & 0.22\\
      RTMDet~\cite{lyu2022rtmdet}  & SW  & 64    & 0.658 & 0.240 & 0.548 & 0.696  & 0.85\\
      RTMDet~\cite{lyu2022rtmdet}  & Zoom    & 92$\dagger$  & 0.622  & 0.248 & 0.636 & 0.704  & 0.41\\
      \rowcolor{LightBlue} 
      RTMDet~\cite{lyu2022rtmdet}  & Ours         & 42$\dagger$  & 0.760  & 0.341 & 0.751 & 0.808 & 3.20\\
      \hline
    \end{tabular}
  % }

  \label{tab:baseline}
\end{table}

% \begin{table}
%   \centering
%   \scalebox{0.9}{
%   \begin{tabular}{@{}l|ccc|ccc@{}}
%     \hline
%     % \toprule
    
%               & AP* & AP$_{50}$ & AP$_{75}$ & AP$_{S}$ & AP$_{M}$ & AP$_{L}$\\
%     % \midrule
%     \hline
%     $\times$ 1   & 0.384 & 0.738 & 0.355 & \textbf{0.132} & \textbf{0.315} & 0.394 \\
%     \hline
%     $\times$1.5  & \textbf{0.480} & \textbf{0.806} & \textbf{0.494} & 0.068 & 0.313 & \textbf{0.509} \\
%     \hline
%     $\times$2    & 0.466 & 0.804 & 0.467 & 0.013 & 0.271 & 0.498 \\
%     % \bottomrule
%     \hline
%   \end{tabular}
%   }
%   \vspace{-0.2cm}
%   \caption{\textbf{Ablation experiments on different expansion factor. } This experiment is based on FasterRCNN. *We set maxdet=1000.}
%   \vspace{-0.5cm}
%   \label{tab:patch_chosse}
% \end{table}

\subsection{Ablation Studies}
\label{sec:ablation}
% We valid our design choices through ablation experiments.
% Experiments about probability boundaries and expansion factors are used to prove the performance improvement brought by related super parameters.
% Different inference strategy demonstrates the double superiority of our method in terms of accuracy and speed.
We valid our design choices through ablation experiments.
Experiments about density boundaries are used to prove the performance improvement brought by related superparameters.
Different inference strategy demonstrates the eight-fold superiority of our method in terms of accuracy and speed.
\\[6pt]
\noindent
\textbf{Density Boundaries}: 
% This probability boundary can be used to further improve the accuracy. 
% This is because if an object has only a small part of its body in the current cell, 
% it will lead to a lower probability. 
% At this time, if the cell is used for fine-grained detection, only half of this object could be detected.
% It is unnecessary to worry that discarding these cells will lose some objects because it will get more accurate results in neighboring cells.
% Although the current stage is region proposal, the recall and accuracy can be evaluated directly on these results. 
% However, we didn't choose this approach because they can't directly reflect the end-to-end performance, 
% so we use a trained FasterRCNN as an evaluation tool and use AP to directly measure the performance of probability boundary.
% The experimental results are shown in Table~\cref{tab:threshold}.
The setting of density boundaries plays a pivotal role in refining the accuracy of our method. If an object is only partially within a grid cell, it results in a lower density score. In such cases, opting for fine-grained detection in these cells may lead to partial detection of the object. However, discarding these cells with lower scores does not equate to losing objects; instead, it often leads to more accurate detections in adjacent cells. To evaluate the impact of density boundaries, we utilized a trained RTMDet model as a proxy, assessing the performance through the Average Precision (AP) metric. The results, as detailed in \cref{tab:threshold}, highlight the effectiveness of this approach. It is important to note that while our values range from 0 to 1, the values for specific areas can exceed 1, as demonstrated in the visualization presented in \cref{fig:model_analysis}. This observation suggests the presence of at least one person in those areas.
\\[6pt]
\noindent
\textbf{Different Inference Strategy}: 
% We designed several different baseline methods, including the naive sliding window and the excellent AutoFocus.
% In order to ensure the fairness of the experiment,  all the detectors are using the same training weights.
% It can be seen from the experimental results in Table~\cref{tab:baseline}, 
% the simplest sliding window also achieves good performance, which is why this method is widely used to deal with large-scale images.
% However, the disadvantages of this method are also obvious. 
% In the experiment based on FasterRCNN, the 16$\times$16 grid strategy can only reach 0.08 FPS, which is far from real-time processing.
% Reducing the number of cells to 8$\times$8 grid can bring higher speed, but it will lead to a sharp drop in accuracy.
% We can conclude that the most common sliding window is an extremely compromised strategy, 
% and we need a better approach in terms of accuracy and speed.
% The experimental results show that we have attained the state-of-the-art performance of all models.
% Especially, MobileNet based SaccadeDet achieves a speed of 6FPS, which is close to the real-time speed and the accuracy is acceptable.
% % The Visualization for results of AutoFocus and our SaccadeDet is shown in  \cref{fig:Auto2TEGA}.
% Our method shows better performance on difficult example because we use a separate density regression model to extract cluster features. 
% AutoFocus uses the same feature to realize object detection and tiny area localization, which is disadvantageous for small and difficult objects.
We compared our method against common baselines, including the straightforward sliding window approach~(SW) and the more sophisticated AutoFocus~(Zoom). All detectors were trained with the same weights to ensure fairness. The sliding window, despite its simplicity, showed commendable performance, elucidating its popularity in handling large-scale images. However, its drawbacks are apparent in terms of speed and accuracy, as shown in ~\cref{tab:baseline}. For instance, a 16$\times$16 grid strategy in Faster R-CNN yielded only 0.08 FPS, far from real-time processing. Opting for an 8$\times$8 grid strategy could enhance the processing speed, however, this comes at the cost of reduced accuracy. Although the AutoFocus method appears to be innovative, further optimization is necessary to effectively adapt it for use in gigapixel-scale scenes.
Our method demonstrated superior performance across all models. This highlights the effectiveness of SaccadeDet, particularly in handling challenging examples, by employing a separate density regression model to extract cluster features, unlike AutoFocus which uses the same features for both object detection and tiny area localization.

\subsection{Application to Whole Slide Imaging}
\begin{figure}[t]
  \centering
     \includegraphics[width=1.0\linewidth]{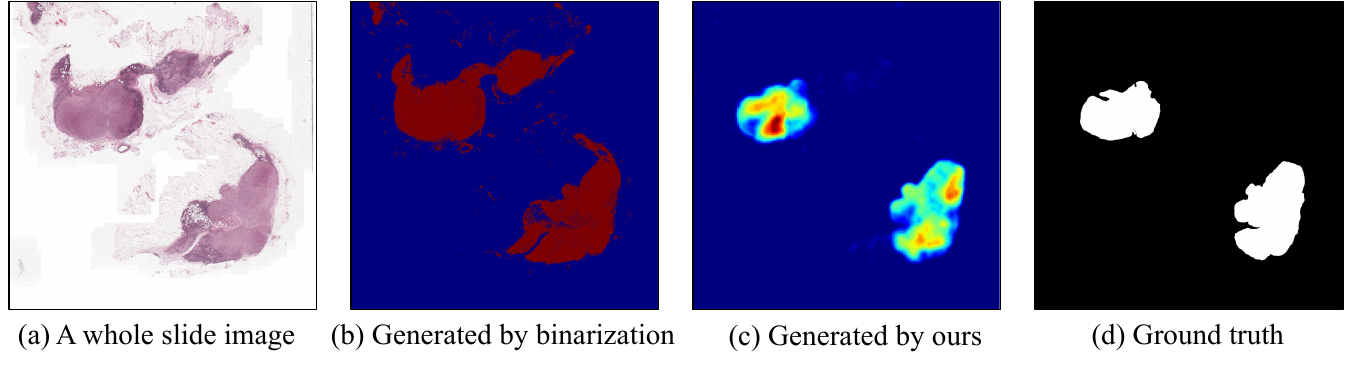}

  \caption{\textbf{Comparison of current preprocessing method and SaccadeDet to generate regions of interest.} SaccadeDet already extracted coarse-grained cancer metastasis regions, while color-based methods can only extract tissue regions.}
  % \vspace{-30pt}
  \label{fig:wsi}
\end{figure}

The practicality of gigapixel imaging, particularly in medical contexts, is evident in Whole Slide Imaging (WSI), where entire microscope slides are digitized into gigapixel files. Our architecture, SaccadeDet, adapts seamlessly to WSI tasks, excelling in various large-scale analyses.
We validated SaccadeDet using the Camelyon16 dataset~\cite{golden2017deep} and Camelyon17 dataset~\cite{bandi2018detection}, containing lymph node tissue slides. Unlike traditional tumor segmentation methods that rely on patch classification at a high resolution, our approach optimizes processing by focusing on areas of interest. Most WSI methods expend resources on processing the extensive background regions. Our architecture counters this inefficiency by applying the saccade method at a downsampled level, significantly narrowing down the regions requiring analysis.
As shown in  \cref{fig:wsi}, SaccadeDet substantially reduces the area needing classification compared to conventional color threshold-based methods. This approach, by pruning non-essential areas at a lower resolution, achieves substantial efficiency gains, with each pixel pruned equating to a significant reduction at the full resolution.

\begin{table}[t]
\centering
\caption{\textbf{Camelyon16 Challenge Results.} Runtime per WSI is reported (unit: minute) in this table.}
\begin{tabular}{@{}lccc@{}}
\toprule
Method & Runtime (WSI) & AUC & FROC \\
\midrule
Harvard \& MIT~\cite{wang2016deep} & 267.2 & 99.4\% & 80.7\% \\
NCRF~\cite{li2018cancer} & 297.2 & - & 81.0\% \\
Fast ScanNet~\cite{lin2019fast} & 8.0 & 98.7\% & 85.3\% \\
Lunit Inc~\cite{lee2018robust} & 454.4 & 98.5\% & 85.5\% \\
B.Kong et al.~\cite{kong2018invasive} & 5.6 & - & 85.6\% \\
LYNA~\cite{liu2019artificial} & 462.0 & 99.3\% & 86.1\% \\
Y.Liu et al.~\cite{liu2017detecting} & 462.0 & 97.7\% & 88.5\% \\
PFA-ScanNet~\cite{zhao2019pfa} & 7.2 & 98.8\% & 90.2\% \\
\midrule
Ours & 2.1 & 99.4\% & 91.1\% \\
\bottomrule
\end{tabular}
\label{tab:cam16}
\end{table}

\begin{table}[t]
\centering
\caption{\textbf{Camelyon17 Challenge Results.} Runtime per WSI is reported (unit: minute) in this table.}
\begin{tabular}{@{}lcc@{}}
\toprule
Team & Runtime (WSI) & Kappa Score \\
\midrule
IMT Inc. (Fast ScanNet) & 8.0 & 0.778 \\
MIL-GPAT & 98.8 & 0.857 \\
VCA-TUe & 464.8 & 0.873 \\
HMS-MGH-CCDS & 26.8 & 0.881 \\
ContextVision & 654.4 & 0.883 \\
Lunit Inc. (2017 winner) & 454.4 & 0.899 \\
DeepBio Inc. & 633.2 & 0.957 \\
PFA-ScanNet~\cite{zhao2019pfa} & 7.2 & 0.905 \\
\midrule
Ours & 2.5 & 0.924 \\
\bottomrule
\end{tabular}
\label{tab:cam17}
\end{table}

\noindent\textbf{Datasets and Evaluation Metrics.} The Camelyon16 dataset comprises 400 WSIs, divided into 270 for training and 130 for testing, and includes lesion-level annotations for all cancerous WSIs. The Camelyon17 dataset consists of 1000 WSIs, with each patient contributing 5 slides (500 for training and 500 for testing), offering pN-stage labels for 100 patients in the training set. Lesion-level annotations are provided for only 50 WSIs in this dataset, specifically including those where Invasive Tumor Cells (ITC) and Micro-metastases are present. To assess slide-level metastasis detection, we employed two metrics from the Camelyon16 Challenge: Area Under the Curve (AUC) and average Free Response Operating Characteristic (FROC), the latter being an average sensitivity calculated at six different false positive rates: 1/4, 1/2, 1, 2, 4, and 8 per WSI. For the classification of pN-stage, the quadratic weighted Cohen’s kappa, as specified in the Camelyon17 Challenge, was utilized as the metric for evaluation.

\cref{tab:cam16,tab:cam17} highlights the superior speed of our method compared to the state-of-the-art methods in the Camelyon16 and Camelyon17 Challenges. It is important to note that our proposed model, SaccadeDet, outpaces previous methods in terms of speed. Furthermore, it not only demonstrates significant accuracy in detection results but also achieves a competitive kappa score.
We believe that our method can be of great help for more applications in the face of more gigapixel scenes in the future.

\section{Conclusion}
\label{sec:conclusion}
% In this paper, we propose a novel gigapixel-level detection framework named SaccadeDet 
% that saccade stage searches the high probability region through multi-scale density regression and the gaze stage focuses on detection with scale normalization. 
% Specifically, we define different sizes of objects regressed on different scales density maps.
% Through the patch proposal method, the density information could be transformed into regional information.
% Then, the patches of different scales will be transformed into standard range by scale normalization and passed to the object detector which is trained under the normalized scale for fine-grained analysis.
% By taking advantage of the density regression on low-resolution images,
% background areas can be fast discarded to achieve efficient inference.
% We have demonstrated our method in tackling large-scale variation and improving speed on a human-centric dataset PANDA.
% Furthermore, an additional experiment on WSI, in which all images are beyond gigapixel, showed our efficiency in pathology. 
% We believe our research will bring new vitality to more research fields.
In this paper, we present SaccadeDet, an innovative dual-stage framework designed for gigapixel-level object detection. SaccadeDet efficiently pinpoints regions of interest via a multi-scale density regression and enhances detection accuracy through scale normalization. We have successfully showcased the effectiveness of our method in addressing large-scale variations and accelerating processing speeds using the PANDA dataset.
When applied to Whole Slide Imaging, SaccadeDet has proven its ability to significantly boost processing speeds. Our results indicate that this framework holds great potential for applications in fields that demand efficient analysis of high-resolution images.

\begin{credits}
\subsubsection{\ackname} This work was supported by National Science and Technology Major Project (No. 2022ZD0119402), "Pioneer" and "Leading Goose" R\&D Program of Zhejiang (No. 2024C01142), National Natural Science Foundation of China (No. U21B2013, 62125106, 62088102, 62322113 and 62376156), Shanghai Municipal Science and Technology Major Project (No. 2021SHZDZX0102), and the Fundamental Research Funds for the Central Universities.

% \subsubsection{\discintname}
% It is now necessary to declare any competing interests or to specifically
% state that the authors have no competing interests. Please place the
% statement with a bold run-in heading in small font size beneath the
% (optional) acknowledgments,
% for example: The authors have no competing interests to declare that are
% relevant to the content of this article. Or: Author A has received research
% grants from Company W. Author B has received a speaker honorarium from
% Company X and owns stock in Company Y. Author C is a member of committee Z.
\end{credits}

\end{document}